\documentclass{icrc2}

\usepackage{times}
\usepackage{graphicx} 

\begin{document}

\title{CYBORG SYSTEMS AS PLATFORMS FOR COMPUTER-VISION ALGORITHM-DEVELOPMENT FOR ASTROBIOLOGY}
\author[1]{Patrick Charles McGuire, Jos\'e Antonio Rodr\'iguez Manfredi, Eduardo Sebasti\'an Mart\'inez, Javier G\'omez Elvira}
\affil[1]{Robotics Laboratory, Centro de Astrobiolog\'ia (CAB), Torrej\'on de Ardoz, Madrid, Spain 28850}

\author[2]{Enrique D\'iaz Mart\'inez, Jens Orm\"o}
\affil[2]{Planetary Geology Laboratory, CAB}

\author[3]{Kai Neuffer, Antonino Giaquinta, Fernando Camps Mart\'inez, Alain Lepinette Malvitte}
\affil[3]{Advanced Computing Laboratory, CAB}

\author[4]{Juan P\'erez Mercader}
\affil[4]{Director, CAB}

\author[5]{Helge Ritter, Markus Oesker, J\"org Ontrup, J\"org Walter}
\affil[5]{Neuroinformatics Group, Computer Science Department, Technische Fakult\"at, University of Bielefeld, P.O.-Box 10 01 31, Bielefeld, Germany 33501 }

\correspondence{ Patrick McGuire: {\it Email:} mcguire@physik.uni-bielefeld.de {\it or} mcguire@inta.es\\ \hspace{5.5cm}{\it Telephone:} +34 91 520 6432 {\it Fax:} +34 91 520 1621}

\firstpage{1}
\pubyear{2003}


\maketitle

\begin{abstract} Employing the allegorical imagery from the film ``The Matrix", we motivate and discuss our `Cyborg Astrobiologist' research program. In this research program, we are using a wearable computer and video camcorder in order to test and train a computer-vision system to be a field-geologist and field-astrobiologist.
\end{abstract}

\section{Matrix Preamble}

{\it We {\bf choose} to use the allegorical symbolism from the film, {\bf The Matrix}} (Wachowski \& Wachowski, 1999).  (The speaker emphasized the {\bf bold-faced} words, and spoke very slowly and with a deep, precise voice, like the Matrix character, Morpheus, did.)

{\it {\bf You} also have a choice. You may {\bf choose} between the blue pill and the red pill} (see Figure 1).
 \begin{figure}[h]
 \includegraphics[width=8cm]{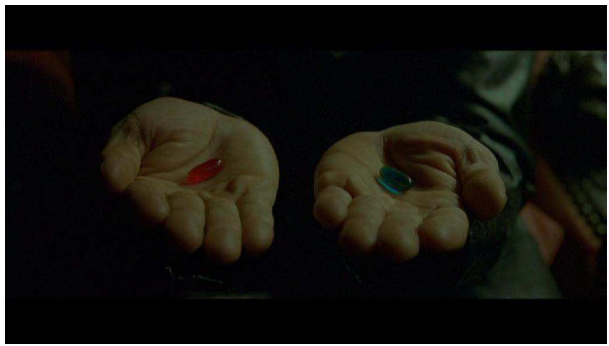}
  \caption{You have a choice between the blue pill on the right and the red pill on the left. (This picture and all the following Matrix pictures
are from the original {\it Matrix} film (Wachowski \& Wachowski, 1999).) }
 \end{figure}

  {\it If you should {\bf choose} the blue pill, the next robots that you send to
Mars to search for life, will have no or little {\bf A.I.} (Artificial
Intelligence) in them. All the science that these robots will do will be
done by astrobiologists and geologists such as yourselves, {\bf here} on the Earth, after the
data has been telemetried by the robots from Mars back to Earth.
  {\bf If} you should {\bf choose} the blue pill, {\bf all} of the intelligence will remain here on the Earth} (see Figure 2).

  {\it {\bf However}, if you should {\bf choose} the {\bf red} pill, these robots, bound for Mars, will have on-board, scientific, astrobiological Artificial
Intelligence. This A.I. will allow {\bf you and} the robots, together, to
accomplish much more science, especially given the several minutes of
delay for delivery of commands from the Earth or for delivery of data from
Mars} (see Figure 2).

 \begin{figure}[h]
 \includegraphics[width=8cm]{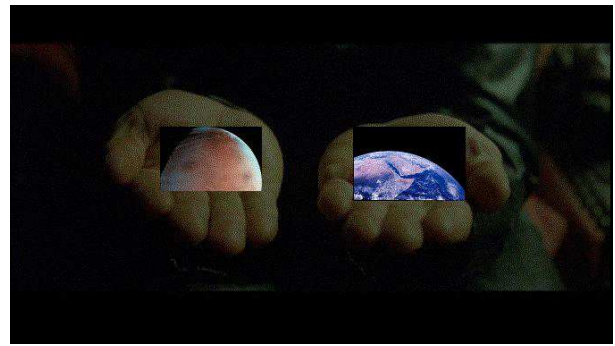}
  \caption{When we explore Mars, you have a choice between keeping all the Geologist intelligence in human form here on the blue planet, the Earth (on the right), and sending significant Geologist intelligence in robotic form to the red planet, Mars (on the left).}
 \end{figure}

  {\it {\bf If} you should {\bf choose} the red pill, {\bf some} of our intelligence will be bound for Mars...}

\section{Live Demonstration of Cyborg Geologist \& Astrobiologist System}
   In the Mars Exploration Workshop in Madrid, we demonstrated to you some of the early capabilities of our `Cyborg' Geologist/Astrobiologist System. We are using this Cyborg system as a platform to develop computer-vision algorithms for recognizing interesting geological and astrobiological features, and for testing these algorithms in the field here on the Earth. 

 \begin{figure}[h]
 \includegraphics[width=4cm]{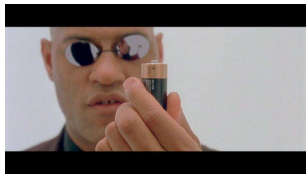}
  \caption{Battery lifetime is important both for running the ruling-machines
in the {\it Matrix} and for reducing the fatigue of the Human Geologist who is 
carrying the wearable computer (and also its batteries) for the Cyborg Astrobiologist project. Battery lifetime was one of the main reasons for choosing a wearable computer from ViA, since this wearable computer contained a high-speed, low-power `Crusoe' CPU.}
 \end{figure}

   One advantage of this Cyborg approach is that it uses human locomotion and human-geologist intuition/intelligence for taking the computer vision-algorithms to the field for teaching and testing, using a `wearable' computer. We concentrate on developing the `scientific' aspects for autonomous discovery of features in computer imagery, as opposed to the more `engineering' aspects of using computer vision to guide the locomotion of a robot through treacherous terrain. This means the development of the scientific vision system for the robot is effectively decoupled from the development of the locomotion system for the robot, which has certain advantages.

   After the maturation of the computer-vision algorithms, we hope to transplant these algorithms from the cyborg computer to the on-board computer of a semi-autonomous robot that will be bound for the red planet.
 \begin{figure}[h]
 \includegraphics[width=8.5cm]{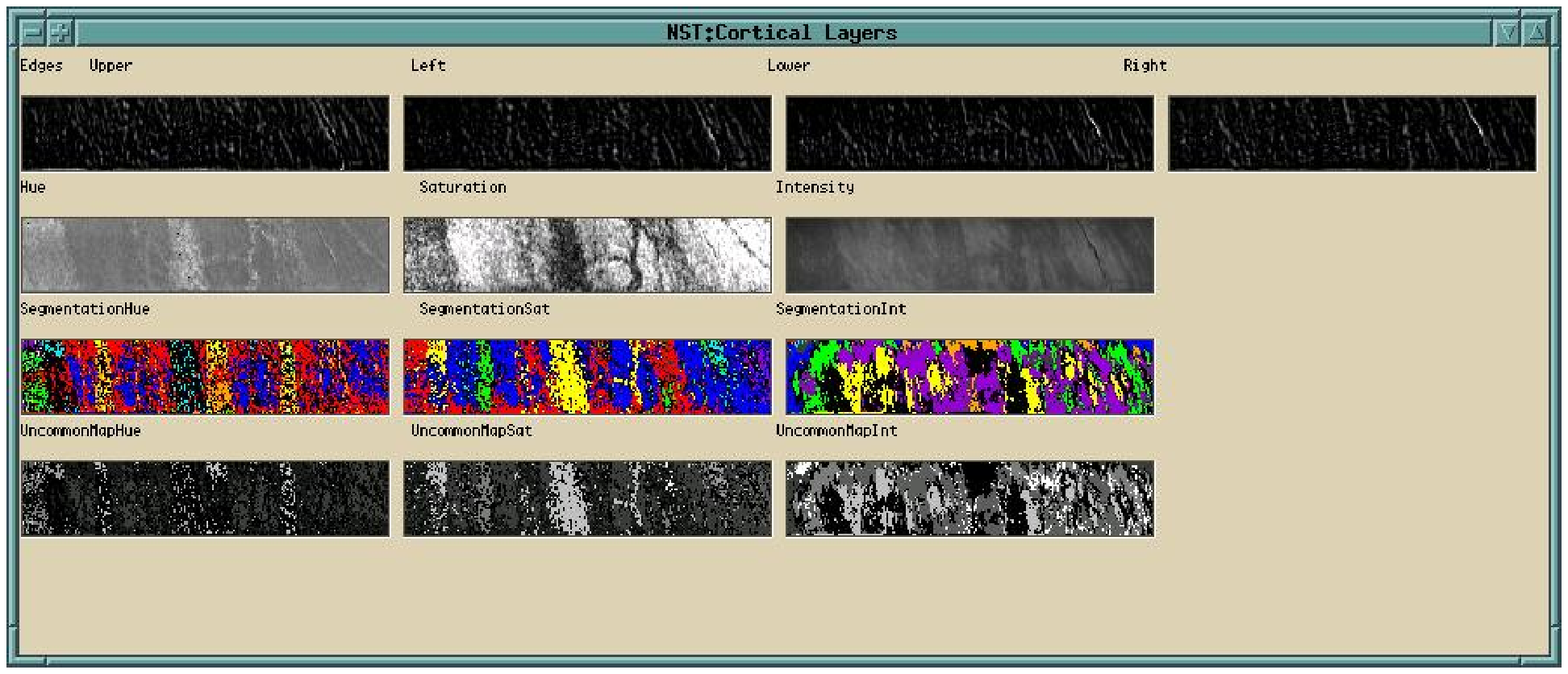}
  \caption{From imagery from the Rio Tinto drilling campaign, we
compute edges in 4 different directions (top row); hue, saturation and intensity (H, S, and I, 2nd row); image-segmentation based upon H,S and I (3rd row); and uncommon maps based upon the segmentation of H, S, and I (4th row).}
 \includegraphics[width=8.5cm]{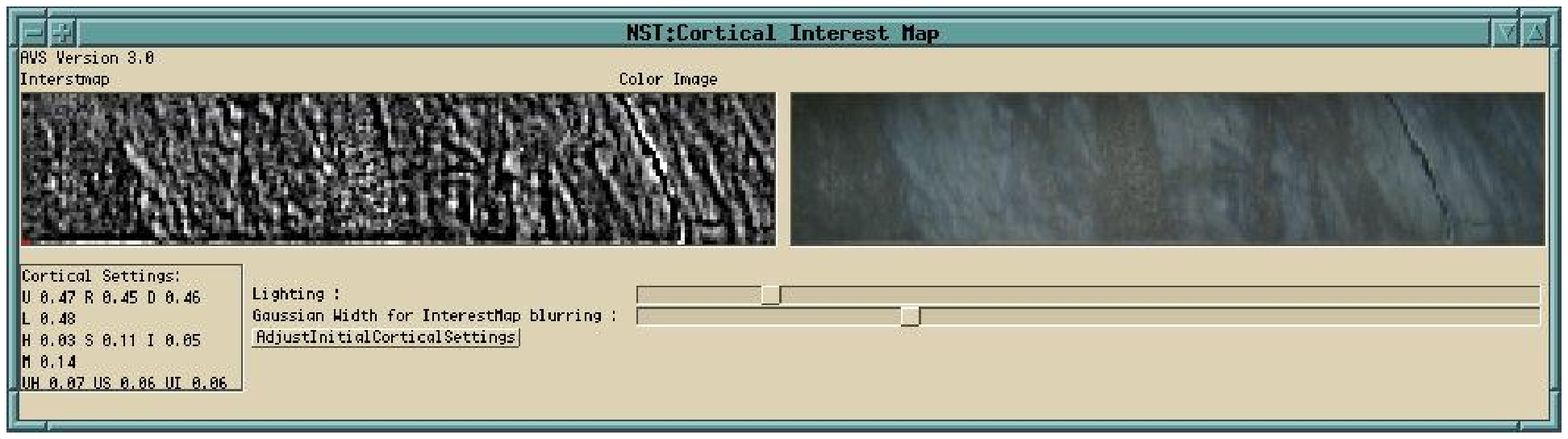}
  \caption{Using the derived images in Figure 4, from the original color image on the right, we add the derived images together with certain coefficients, in order to determine the cortical interest map, shown on the left. The coefficients automatically adapt, depending on the images already processed.}
 \end{figure}

The digital hardware of the Cyborg Astrobiologist system consists of a 667 MHz wearable computer (from ViA Computer Systems in Minnesota) with a `power-saving' Transmeta `Crusoe' CPU, a Head Mounted Display (from Tekgear in Virginia, via the Spanish supplier Decom in Val\`encia) with native pixel dimensions of 640 by 480 that works well in bright sunlight, thumb operated hand mouse, and a SONY `Handycam' color video camera. The power-saving aspect of the Crusoe processor is important because it extends battery life (Figure 3), meaning that the human does not need to carry very many spare batteries, meaning less fatigue for the human geologist. A single battery, which weighs about 1kg, can last for 3 hours or so for this application.

   During the live demonstration of the Cyborg Astrobiologist system
we put a large poster in the front of the lecture hall of the mosaic image from the Mars Pathfinder site. The lecturer disconnected the VGA output of the wearable computer from the TekGear Head-Mounted Display (HMD), and then  reconnected the VGA output of the ViA wearable computer to the LCD projection system of the auditorium. A SONY Handycam provided real-time imagery of the poster of the Mars Pathfinder site to the wearable computer via an IEEE1394/Firewire communication cable. The wearable computer processed the images, to compute a map of interesting areas (Figure 5), at a rate of about 1 image per 8 seconds. This is rather slow, but considering that the system was doing a lot of computations, it certainly is not unreasonably slow. The computations included: two-point intensity-correlations for image-segmentation (Figure 4, \, Haralick, Shanmugan \& Dinstein, 1973; Haddon \& Boyce, 1990), edge-computation using Sobel operators (Figure 4), and adaptive fusion of 9-11 different derived images computed from the original color image (Figure 4) into a single `cortical' interest map (Figure 5, Fislage, Rae \& Ritter, 2000).  We also used the image-segmentation map to compute an `uncommon' map, which gives highest weight to those regions of smallest area (Figure 4).

 \begin{figure}[h]
 \includegraphics[width=6cm]{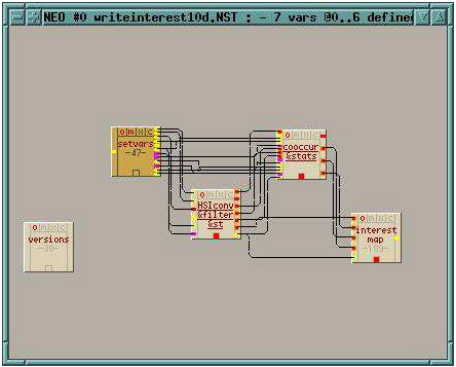}
 \includegraphics[width=5cm]{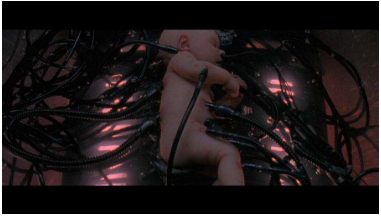}
  \caption{A NEO program, and the {\it Matrix} character Neo as an infant. The 
wires serve as inputs and outputs both to Neo and to the NEO icon sub-programs.}
 \end{figure}

   We tried to point out during the demonstration how the system, when pointed at a poster from the Mars Pathfinder site, was able to find certain areas of the images interesting, for example the edges of the hill on the horizon, the shadow under one of the nearby rocks, or some of the rocks that had more saturated colors than the rest of the image.

\section{We believe in NEO}

 \begin{figure}[h]
 \includegraphics[width=8cm]{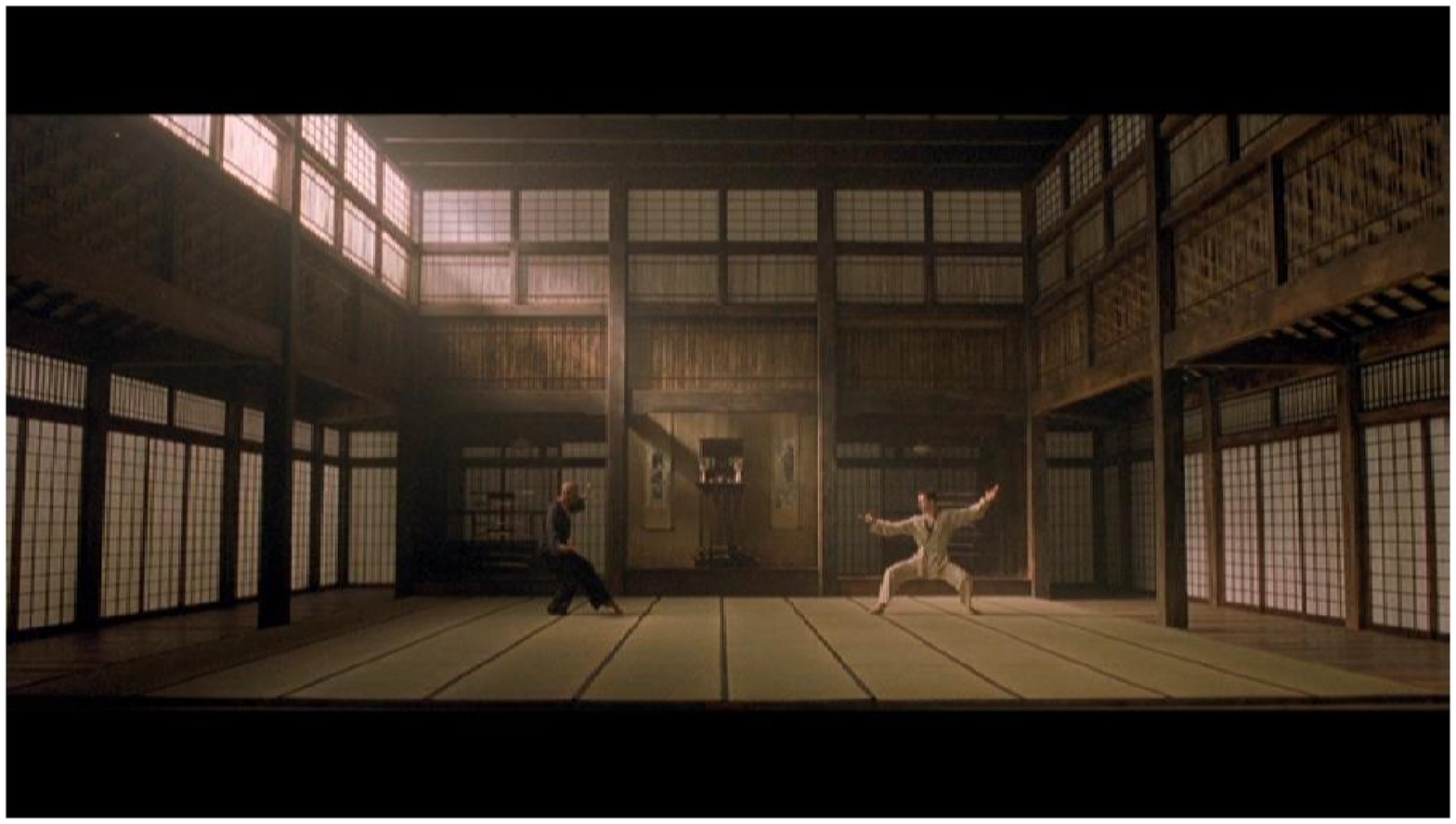}
  \caption{Cyborg geologist training program. For symbolic purposes, we 
refer to the man on the left (Morpheus) to be the Human Geologist, and the man on the right (Neo) to be the computer-vision algorithm, which is currently being trained by the Human Geologist.}
 \includegraphics[width=4cm]{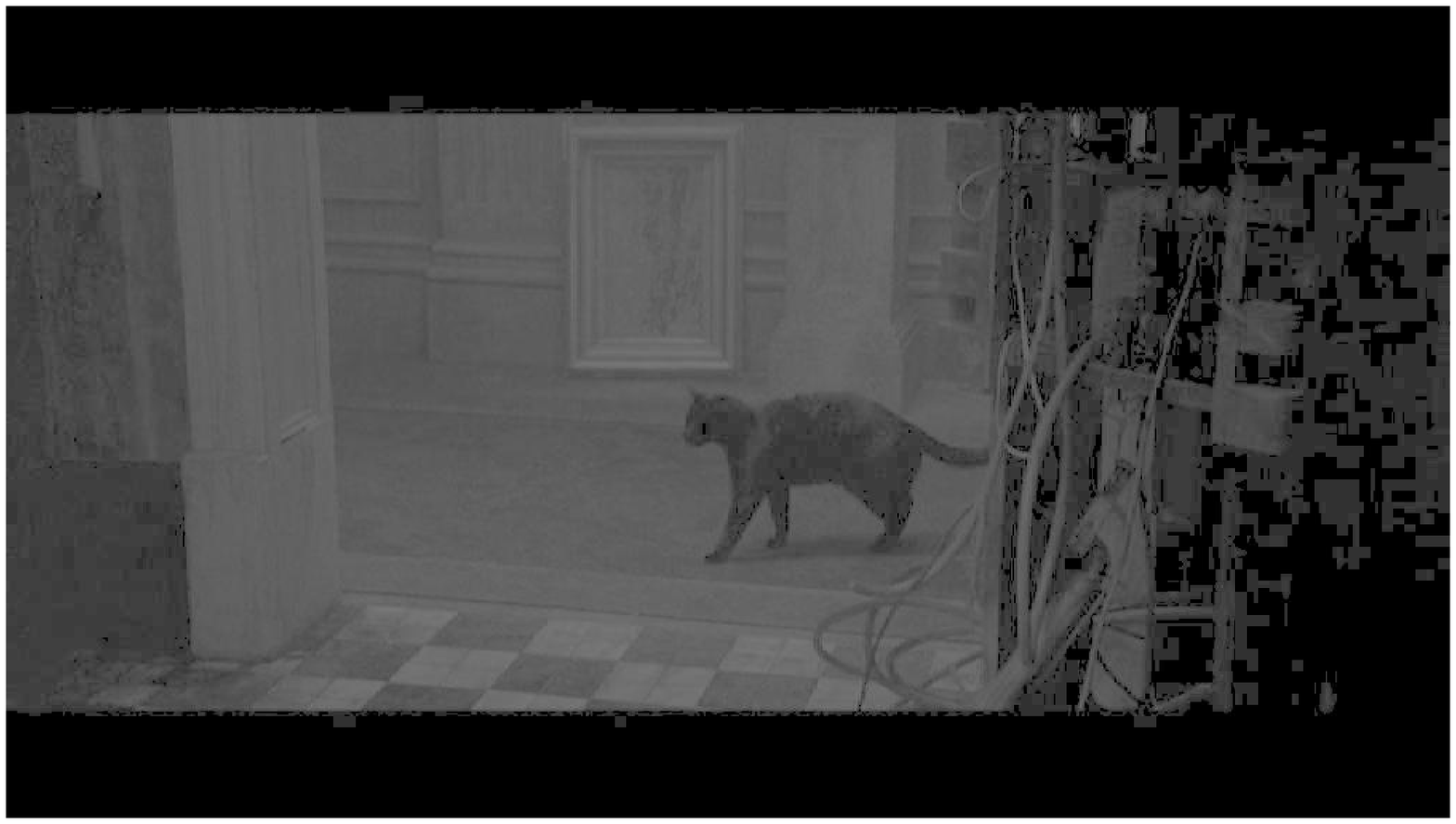}
  \caption{Part of the Matrix program which is repeated a second later (a deja vu event), and which signified to the team that the programmer was making change to the program, to correct a perceived flaw. This ability to 
make debugging changes and view their results without restarting the whole program is one of the key capabilities of the NEO graphical-programming language.}
 \end{figure}

For programming the Cyborg Astrobiologist Computer system, we use 
a Graphical Programming language, called NEO, from the University of Bielefeld (Germany), (Ritter et al 2000). The `Neural EditOr' (NEO) has such capabilities that the programmer can write code rather rapidly and visually, by removing most of the tedious aspects out of programming. It does this by replacing many keystrokes with several mouse clicks, and by graphically showing the program and its flow. A subroutine is replaced by a graphical icon with input and
output pins of varying types. And different icons can be connected by wires
to pass data between two icons, from the output of one to the input of the next (see Figure 6). A NEO icon can be: 1) a container containing graphical NEO subcircuits, 2) a short C or C++ subroutine written by the user and `interpreted' by NEO, or 3) a more complex C or C++ subroutine that is precompiled for speed enhancement.

 \begin{figure}[h]
 \includegraphics[width=7cm]{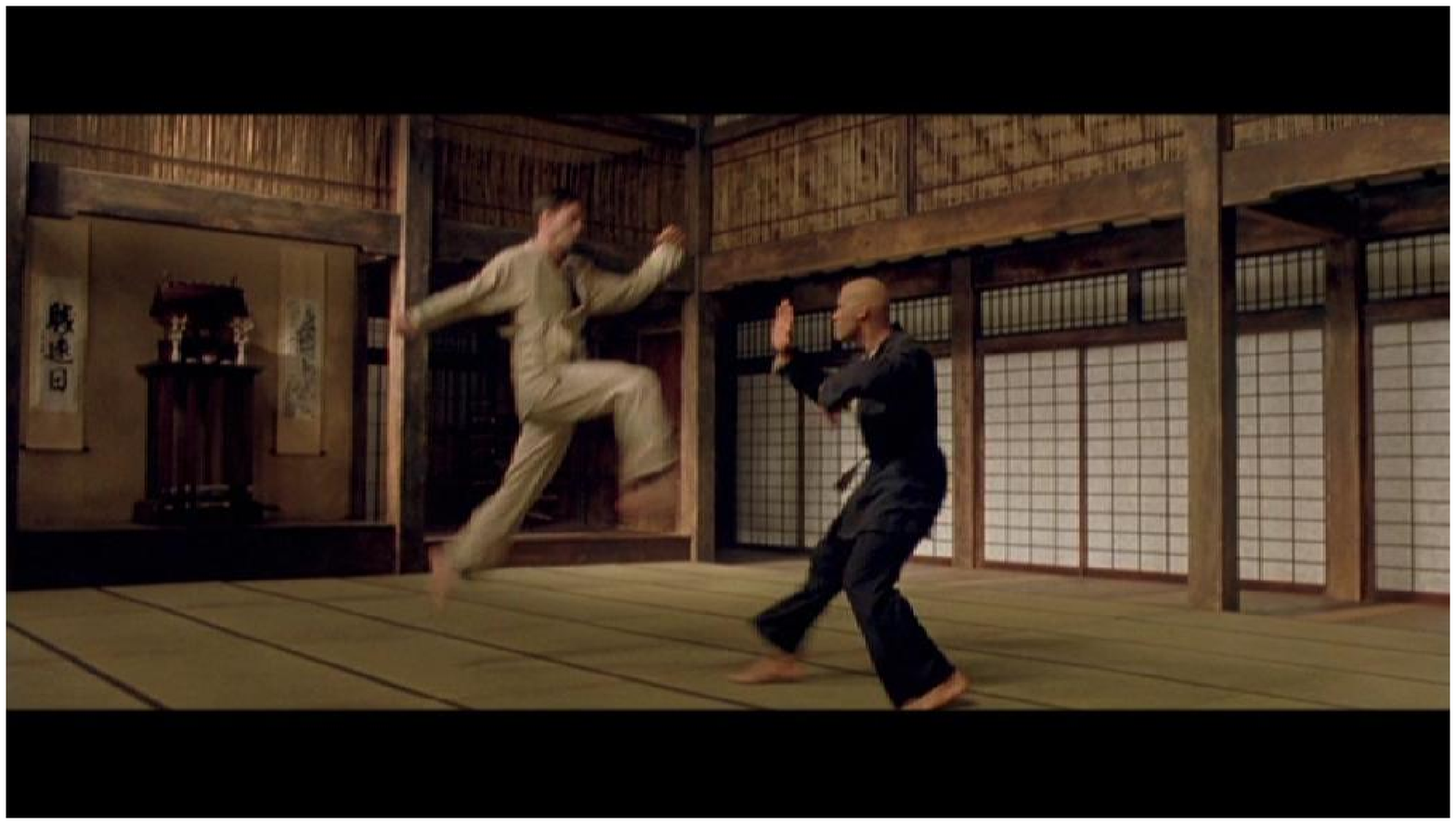}
  \caption{Cyborg Geologist beginning to learn from the Human Geologist. }
 \includegraphics[width=7cm]{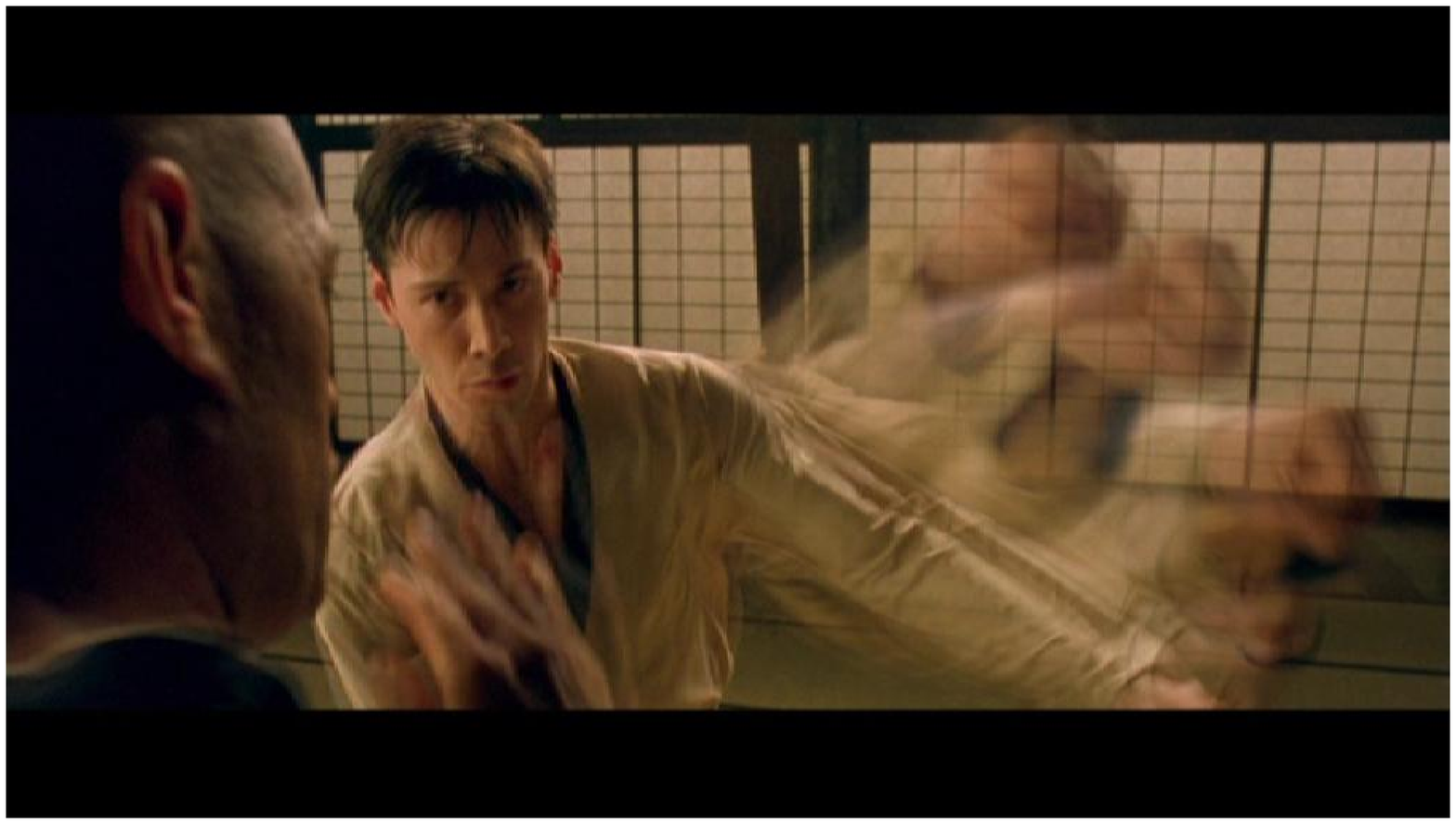}
  \caption{Cyborg Geologist beginning to defeat the Human Geologist. }
 \includegraphics[width=7cm]{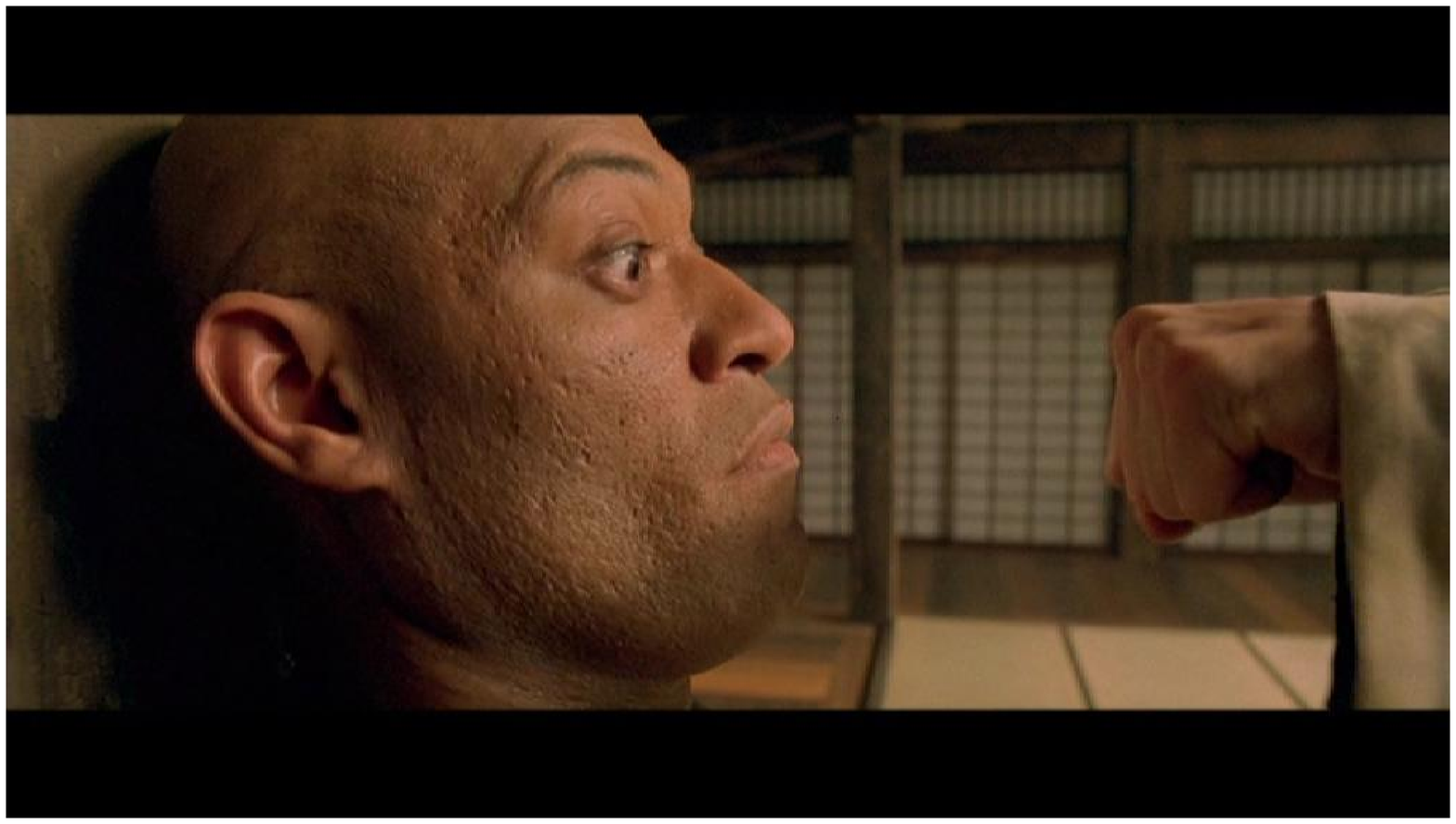}
  \caption{Cyborg Geologist beating Human Geologist. }
 \end{figure}

 \begin{figure}[h]
 \includegraphics[width=5cm]{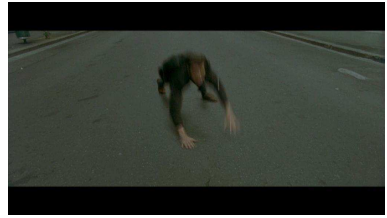}
  \caption{Cyborg Geologist intelligence trying to replace the intelligence
of a Human Geologist before the Cyborg Geologist intelligence is completely ready, causing a fall when it tries to make its first jump from Earth to Mars. }
 \end{figure}

   One example of the power of NEO is that the whole program does not need to
be run; a single icon can be executed or modified without running the rest of the program. Therefore, if the program is misbehaving and doing things that the Human-Geologist trainer of the NEO circuit finds to be inadequate (see Figure 7 for a symbolic setting of a Human Geologist training a NEO Cyborg Geologist circuit), then the NEO
 circuit can then be stopped after a certain number of executions of the whole
 circuit, and one or more of the  individual NEO icons (sub-circuits) can then
 be executed without running the whole program. This is especially useful since the inputs to these NEO sub-circuits are frozen; hence, if a bug exists, the user can rerun the faulty sub-circuits many times, and the user can even make changes to and eventually correct the faulty NEO sub-circuit, without recompiling the whole program and even keeping the inputs frozen (recall the `deja vu' black-cat events in the Matrix Movie, as in Figure 8).

 Eventually, and rather optimistically, the ability of the Human Programmer or the Human Geologist to find and correct faults in the performance of the NEO Cyborg Geologist circuit will diminish (see Figure 9). This improvement in the abilities of the NEO Geologist program may even someday allow the Cyborg Geologist to surpass the abilities of the original Human Geologist trainer (see Figures 10 and 11).

But more realistically, much work and preparation needs to be completed
before the first mostly-autonomous systems can be trusted to go to Mars
(Figure 12).

\section{Outlook}
   The NEO programming for this Cyborg Geologist project was initiated
with the SONY Handycam in April 2002. The wearable computer arrived in June 2003, and the head mount\-ed display arrived in November 2003. We have a functioning human and hardware and software Cyborg Geologist system now. We have even extended the system somewhat by connecting the serial port of the wearable computer to a Pan Tilt mount, so that the camera can be pointed in different directions, automatically by the computer. 

With the functioning Cyborg Geologist system, we plan extensive field tests in early 2004. The Cyborg Geologist system will test the basic capabilities of our image-segment\-ation and interest-map algorithms on the grounds of our Spanish research institute, which is surrounded by Spanish prairie with a small lake nearby. And in the same time frame, the Cyborg Geologist system will accompany a field geologist to sites of more geological interest (i.e. layered rock outcrops in the Castillian or Guadalajaran mountains) for more advanced testing of these computer-vision algorithms.

\section{Query after the talk}
{\it Query:} Do you think a system
could be developed that could properly take into account
the judgements of the ro\-bots, astronauts, ground-controllers and administrators, when it comes to making decisions in the future exploration of Mars?
{\it (asked
by a dark-haired, bespectacled gentleman in the right half of the audience; our apologies if his question was misinterpreted.)}

{\it Response:} Hmmmm. The answer to your question is another
 question: {\it (answered seriously, but in terms of the symbolism and language of the two characters, the Oracle and the Architect, in the {\it Matrix} film)} ``Would you really want to implement such a system?" {\it pause}

In order to do that, you would need an `Architect'. {\it pause} 

There is a technique {\it (answered more
seriously, and without intentional symbolism from the lecturer)}
in Artificial Intelligence programming that could be used for this application, called ``Modular Neural Networks" (cf. Haykin 1990), wherein the judgements of individual subprogram modules (i.e. the individual judgements of a robot, an astronaut, a ground-controller or a NASA administrator) are all combined into one master judgement. The Modular Neural Network learns from past mistakes, and
its errors are passed down to correct the `master weights' (of the connections from judgements of the individual modules to the master judgement), and also
to correct the internal workings of the the individual modules themselves.



\begin{acknowledgements}
P. McGuire would like to thank the Ramon y Cajal Fellowship program in Spain,
as well as certain individuals for assistance or conversations:
Mar\'ia Paz Zorzano Mier, Josefina Torres Redondo, V\'ictor R. Ruiz, Julio Jos\'e Romeral Planell\'o, Gemma Delicado, Jes\'us Mart\'inez Fr\'ias, Susanna Schneider, Gloria Gallego, Carmen Gonz\'alez, Ramon Fern\'andez,  Colonel Santamaria, Carol Stoker, Paula Grunthaner, Fernando Ayll\'on Quevedo, Javier Mart\'in Soler, Robert Rae, Claudia Noelker, and Jonathan Lunine. The equipment used in this work was purchased by grants to our
Center for Astrobiology from its sponsoring research organizations, CSIC and INTA.
\end{acknowledgements}

\end{document}